\documentclass[a4paper, 10pt, conference]{ieeeconf}     \IEEEoverridecommandlockouts                  
\usepackage{graphicx,amsmath,hyperref,color,tikz,pgfplots,subfig,booktabs,multirow,array,multirow,siunitx,pgfplotstable,filecontents,scalerel}
\pgfplotsset{compat=1.18}
\usetikzlibrary{svg.path}
\definecolor{orcidlogocol}{HTML}{A6CE39}
\tikzset{
  orcidlogo/.pic={
    \fill[orcidlogocol] svg{M256,128c0,70.7-57.3,128-128,128C57.3,256,0,198.7,0,128C0,57.3,57.3,0,128,0C198.7,0,256,57.3,256,128z};
    \fill[white] svg{M86.3,186.2H70.9V79.1h15.4v48.4V186.2z}
                 svg{M108.9,79.1h41.6c39.6,0,57,28.3,57,53.6c0,27.5-21.5,53.6-56.8,53.6h-41.8V79.1z M124.3,172.4h24.5c34.9,0,42.9-26.5,42.9-39.7c0-21.5-13.7-39.7-43.7-39.7h-23.7V172.4z}
                 svg{M88.7,56.8c0,5.5-4.5,10.1-10.1,10.1c-5.6,0-10.1-4.6-10.1-10.1c0-5.6,4.5-10.1,10.1-10.1C84.2,46.7,88.7,51.3,88.7,56.8z};
  }
}
\newcommand\orcidicon[1]{\href{https://orcid.org/#1}{\mbox{\scalerel*{
\begin{tikzpicture}[yscale=-1,transform shape]
\pic{orcidlogo};
\end{tikzpicture}
}{|}}}}
\usepackage{hyperref}
\usepackage[T1]{fontenc}
\usepgfplotslibrary{colorbrewer}
\usetikzlibrary{pgfplots.statistics, pgfplots.colorbrewer}

\title{\LARGE \bf Experimental Evaluation of Road-Crossing Decisions by Autonomous Wheelchairs against Environmental Factors}
\author{
\authorblockN{Franca Corradini\orcidicon{0009-0009-5867-9478}
, Carlo Grigioni\orcidicon{0009-0002-0794-0612},
 Alessandro Antonucci\orcidicon{0000-0001-7915-2768},
 J\'er\^ome Guzzi\orcidicon{0000-0002-1263-4110}
and Francesco Flammini\orcidicon{0000-0002-2833-7196}}
\authorblockA{
IDSIA USI-SUPSI\\University of Applied Sciences and Arts of Southern Switzerland\\Lugano (Switzerland)\\ name.surname@supsi.ch}}

\begin{document}

\maketitle

\begin{abstract}

Safe road crossing by autonomous wheelchairs can be affected by several environmental factors such as adverse weather conditions influencing the accuracy of artificial vision. Previous studies have addressed experimental evaluation of multi-sensor information fusion to support road-crossing decisions in autonomous wheelchairs. In this study, we focus on the fine-tuning of tracking performance and on its experimental evaluation against outdoor environmental factors such as fog, rain, darkness, etc. It is rather intuitive that those factors can negatively affect the tracking performance; therefore our aim is to provide an approach to quantify their effects in the reference scenario, in order to detect conditions of unacceptable accuracy. In those cases, warnings can be issued and system can be possibly reconfigured to reduce the reputation of less accurate sensors, and thus improve overall safety. Critical situations can be detected by the main sensors or by additional sensors, e.g., light sensors, rain sensors, etc. Results have been achieved by using an available laboratory dataset and by applying appropriate software filters; they show that the approach can be adopted to evaluate video tracking and event detection robustness against outdoor environmental factors in relevant operational scenarios.
\end{abstract}

\section{Introduction}

Safe road crossing is a paramount concern for autonomous wheelchair technology, particularly in the face of environmental challenges like adverse weather conditions. Previous studies have explored the fusion of multi-sensor information to support road-crossing decisions in cooperative robotic systems made of \textit{Autonomous Wheelchairs} (AWs) and flying drones \cite{grigioni2024safe}. However, the impact of specific environmental factors on obstacle tracking performance by artificial vision remains a critical area requiring deeper investigation. In this study, we aim to address this gap by focusing on fine-tuning obstacle tracking performance and evaluating its robustness against outdoor environmental factors such as fog, rain, and darkness. By quantifying these effects, we seek to identify thresholds where tracking accuracy falls below acceptable levels, facilitating timely warnings and sensor reconfigurations to enhance overall safety as required by international standards \cite{tas23, Flammini2020}. Additionally, the integration of supplementary sensors allows for improved detection of critical situations. Through our research, we contribute to advancing the trustworthiness \cite{9979717} of AW navigation systems, ultimately empowering individuals with mobility impairments to navigate their environments with greater confidence and independence.


In order to evaluate performance for both single sensors and sensor fusion, we apply appropriate environmental filters to the output coming from the cameras, before applying a YOLO model fine-tuned to our specific case.

The main original contributions of this paper are summarized as follows:

1) We fine tune artificial vision performance compared to previous studies in order to improve video tracking results by combining diverse modules and training datasets, including generic and specific ones.

2) We experimentally evaluate the performance in video tracking and estimation of road crossing danger function after simulating adverse environmental conditions. Those are generated by using appropriate filters applied to a video dataset built in laboratory environment, which is equipped with wheeled robots, sensors, and optical tracking infrastructure to build the so called ground-truth.

The research presented in this paper has been performed in the context of an international research project named REXASI-PRO (REliable eXplAinable Swarm Intelligence for People with Reduced mObility)\footnote{\url{https://rexasi-pro.spindoxlabs.com}.}, aimed at introducing an innovative engineering framework for safe navigation of autonomous wheelchairs using trustworthy artificial intelligence.

The remaining of this paper is structured as follows. Section~\ref{sec:background} provides background information about the road crossing danger function, the laboratory dataset used for the experimentation, and the sensor fusion approach. Section~\ref{sec:method} presents the evaluation methodology, the tools, and the laboratory setup used for the experiments. Section~\ref{sec:experiments} provides and discusses the results of the experimentation summarizing the main findings. Finally, Section~\ref{sec:conc} provides conclusions and hints for future developments.

\section{Background}\label{sec:background}
In urban vehicular traffic environment, pedestrian and wheelchair users are among the most exposed subjects \cite{Pedestrian,Obstacles_wheel}. The road-crossing scenario is therefore an essential situation to analyse in terms of pedestrian–vehicle conflict to prevent accidents \cite{ELHAMDANI2020102856}. In this context, wheelchair users may encounter further disadvantages, due to a possible lack of reaction, or inability to analyze the danger. 

In reference \cite{grigioni2024safe}, we proposed a danger evaluation approach for road crossing using a sensor system based on multiple, diverse, and redundant components. Sensor fusion was applied at different levels of information processing and to test the method we created a novel dataset recorded in our laboratory using a simplified scenario, where laboratory devices were used to simulate the crossing scenario agents. Part of the dataset is used for testing as a case of study. The results highlighted the advantages of using diverse sensors to take safer road crossing decisions, especially when information fusion is applied at the lowest level of data processing. 


\begin{figure}[htp!]
    \centering
    \includegraphics[width=8cm]{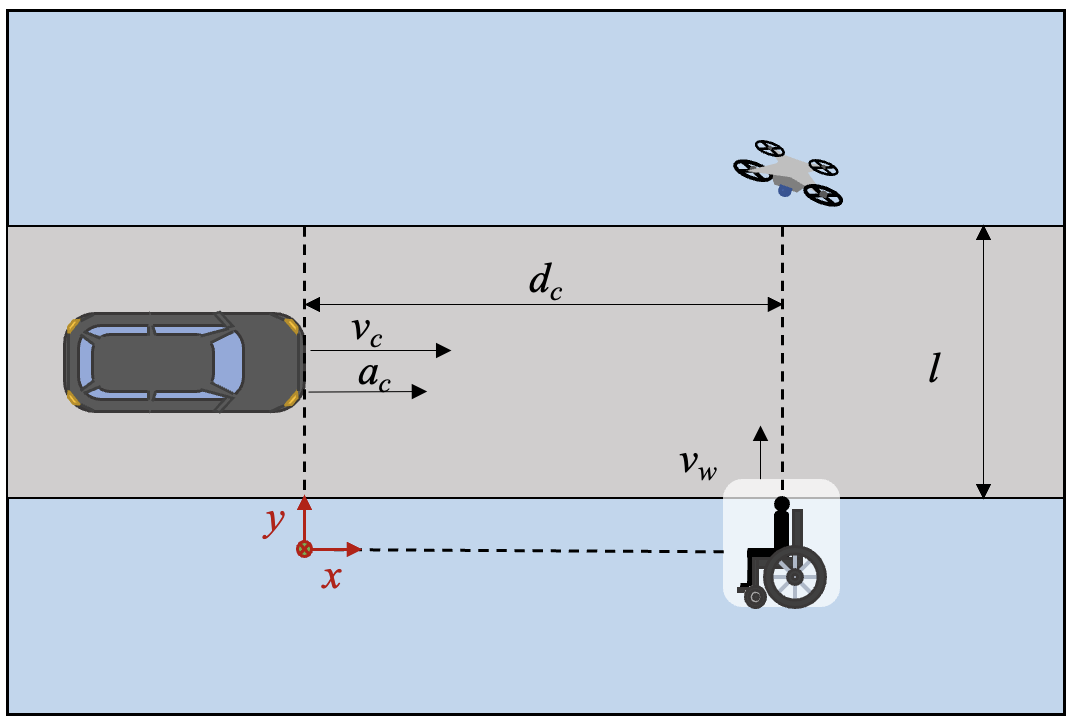}
    \caption{Kinematics of the danger function design.}
    \label{fig:raod_crossing}
\end{figure}

\subsection{A danger function to support road-crossing decisions}\label{sec:df}
Based on the scope of the project where this experimentation has been performed, the reference system is composed by an AW and a drone providing an additional perspective over oncoming vehicles. The reference scenario is depicted in Fig.~\ref{fig:raod_crossing}.

We used a continuous function $g$, called \emph{Danger Function} (DF) reflecting a kinematic analysis of the road-crossing scenario at a particular instant of time $t$. For decision-making, we used the output provided by the DF: the crossing is considered dangerous if the corresponding DF values exceed a given threshold $g^*=1$. The DF is defined as:
\begin{equation}\label{eq:complete}
g(d_c, v_c, a_c) := \frac{h({v}_c) + k \cdot f(a_c)}{\log (d_c + \epsilon)}\,,
\end{equation}
where $d_c$, $v_c$ and $a_c$ are respectively the distance, relative speed and relative acceleration between the vehicle and the pedestrian. $k=0.1$ and $\epsilon=0.6$ are set following heuristics, while $h$ and $f$ are linear transformations with threshold applied on the components $v_c$ and $a_c$. The linear transformation of the speed is
\begin{equation}\label{eq:speed_component}
h(v_c) :=
\begin{cases} 
0&\text{if }v_c \leq \underline{v}_c\,,\\
\frac{v_c-\underline{v}_c}{\overline{v}_c-\underline{v}_c}&\text{if }\underline{v}_c< v_c\leq \overline{v}_c\,,\\
1&\text{if }v_c > \overline{v}_c\,,\\
\end{cases}
\end{equation}
and, for the acceleration
\begin{equation}\label{eq:acceleration_component}
f(a_c):=
\begin{cases} 
-1&\text{if }a_c\leq -\overline{a}_c\\
\frac{a_c+\underline{a}_c}{\overline{a}_c-\underline{a}_c}&\text{if }-\overline{a}_c < a_c \leq -\underline{a}_c\\
0&\text{if }-\underline{a}_c<a_c\leq \underline{a}_c\\
\frac{a_c-\underline{a}}{\overline{a}_c-\underline{a}_c}&\text{if }\underline{a}_c < a_c  \leq \overline{a}_c\\
1 & \text{if }a_c > \overline{a}_c\,.\\
\end{cases}
\end{equation}

where $\underline{v}_c$ = $0.05 m/s$, $\overline{v}_c= 0.65m/s$, $\underline{a}_c=1m^2/s$, $\overline{a}_c=10m^2/s$.




\subsection{A laboratory dataset for simulated road-crossing}\label{sec:data_gen }

\begin{figure}[htp!]
    \centering
    \includegraphics[width=8cm]{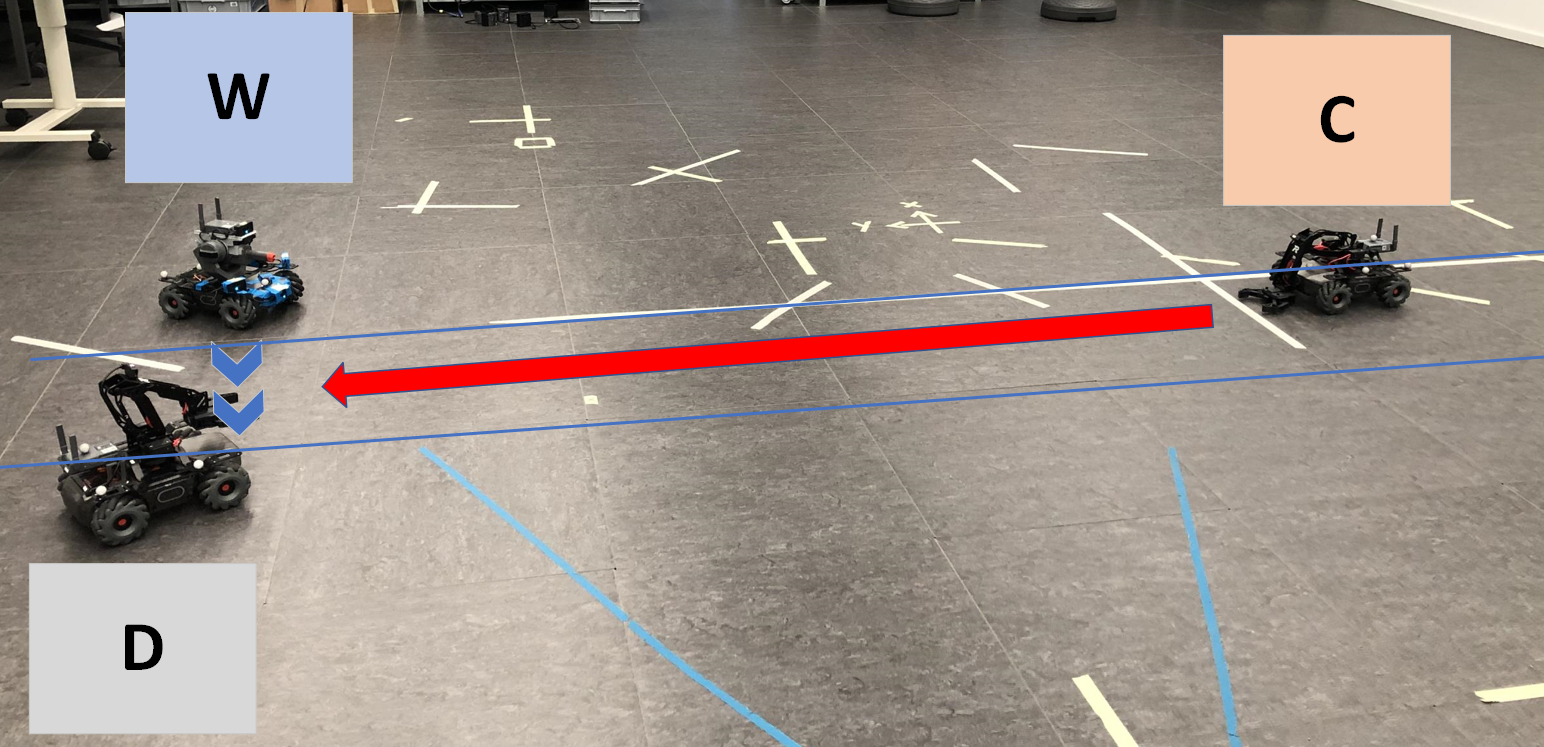}
    \caption{A data collection scenario: the trajectory of RM$_{\mathrm{c}}$ is in red and the crossing path of RM$_{\mathrm{w}}$ in blue.} 
    \label{fig:Scenes}
\end{figure}

In order to build a dataset for the reference scenario, we used three wheeled ground robots named \textit{Robomasters}\footnote{\url{https://www.dji.com/ch/robomaster-ep}.} (RMs) equipped with cameras and distance sensors. This allows to evaluate system performance in the simplified experimental setup depicted in Fig.~\ref{fig:Scenes}. A dataset has been recorded in the \emph{IDSIA Autonomous Robotics Laboratory}.\footnote{\url{https://idsia-robotics.github.io}.} We shared the dataset with the scientific community by making it available on a publicly accessible repository.\footnote{\url{https://huggingface.co/datasets/carlogrigioni/safe-road-crossing-aw-dataset}.} We collected ground-truth poses from a very accurate motion tracking system, which were used to evaluate the performance of the system from a quantitative point of view.

For each experimental run, we recorded data from two cameras and a set of range sensors located on the two RMs acting as the AW (RM$_{\mathrm{w}}$) and the drone (RM$_{\mathrm{d}}$), and the motion tracker. Therefore, data from cameras were elaborated to obtain bounding boxes of the RM performing as the vehicle (RM$_{\mathrm{c}}$) in each frame of the recorded videos. 

Overall, fifteen runs were recorded for two different setups (scenario B is depicted in Fig.~\ref{fig:Scenes}), which differ mainly in the background of the videos recorded and the length of the route of RM$_{\mathrm{v}}$. The experiments of the first setup (scenario A) last approximately \SI{6}{s}. In scenario B the initial distance of the RM$_{\mathrm{c}}$ with respect to the RM$_{\mathrm{w}}$ is longer, allowing for more complex patterns in movement of the obstacle during experiments for a duration of approximately \SI{9}{s}.


Additional information on the devices and the laboratory can be found in reference \cite{grigioni2024safe}, and in the associated repository.\footnote{\url{https://github.com/CarloGrigioni/safe_roadcrossing_aw?tab=readme-ov-file}.}


\subsection{Sensor fusion}\label{sec:fusion}
When sensors data are fused, among the various aspects to consider, an important one is to define the default value when the obstacle is not detected. This may be due either to the actual absence of the obstacle or to the inability of the sensor to perceive it.
In \cite{grigioni2024safe} only a partial discussion is made, considering the contribution of each sensor absent until the first obstacle detection. At this point, when the sensor no longer sees the obstacle, the last value detected is kept constant. This choice proved to be inadequate for the data coming from the cameras, which do not have an automatic reset procedure, such as the range sensors.

Information fusion from sensors and cameras can be done at different levels. We can directly average the distances measured by the sensors or perform the fusion only after the evaluation of the DF for each sensor or even at the level of the binary road-crossing decisions (e.g., by a voting procedure). 

Here we focus on the fusion of the distances, which, according to \cite{grigioni2024safe}, is the one providing best performances when a trivial average of the sensors data is applied without considering whether the sensor detects an obstacle or not, as described in the Subsec.~\ref{sec:data_gen }.

\section{Evaluation methodology}\label{sec:method}
Since the major goal of this paper is to study how environmental (i.e., exogenous) factors affect road-crossing decisions, we need to simulate relevant outdoor and environmental disturbances by means of appropriate \emph{filters}. Those filters, applied to the frames recorded by the video cameras, can mimic the presence of adverse conditions such as darkness, rain and fog.

\begin{figure}[htp!]
\centering
\begin{minipage}{1\linewidth}
\centering    \subfloat{\includegraphics[width=.32\linewidth]{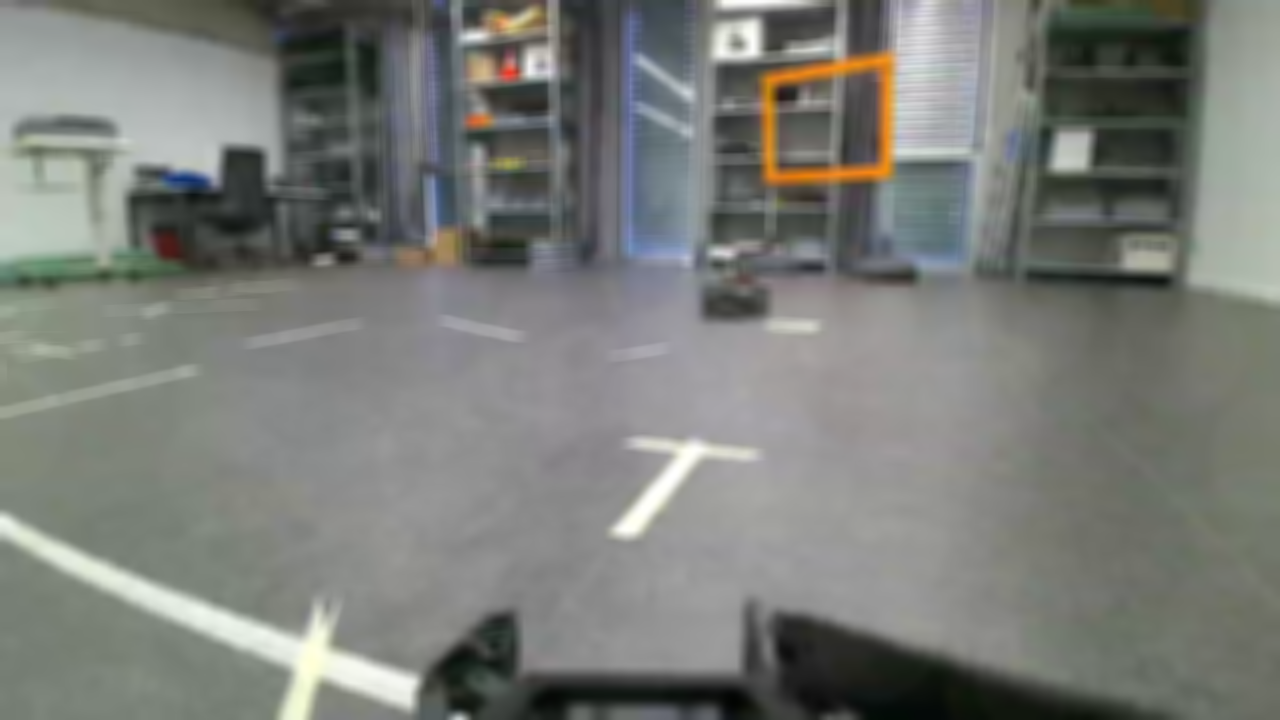}}
\subfloat{\includegraphics[width=.32\linewidth]{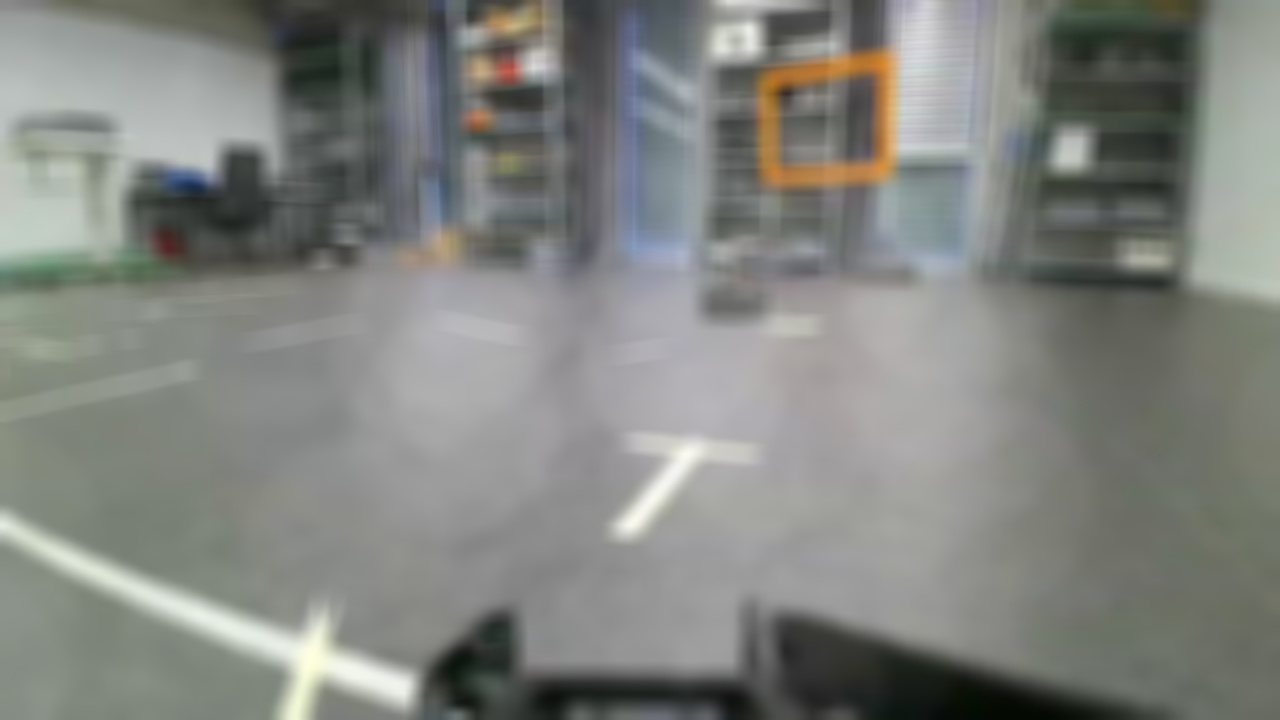}}    \subfloat{\includegraphics[width=.32\linewidth]{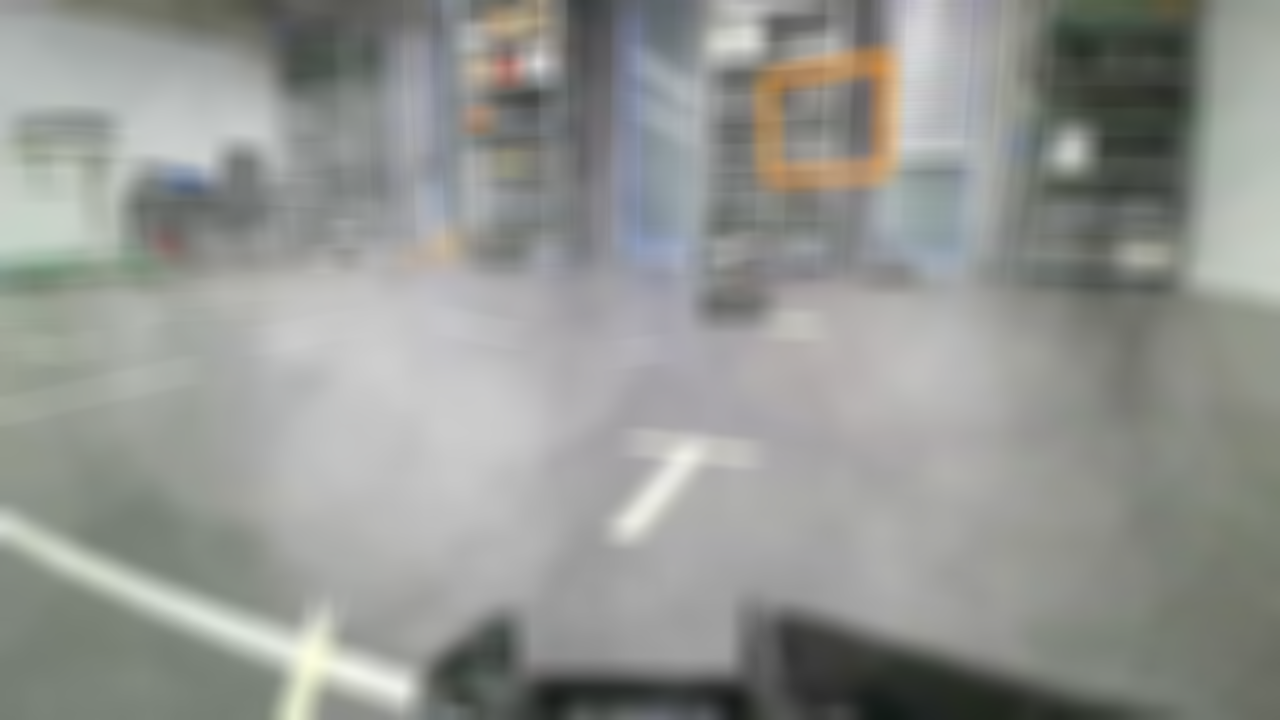}}
\caption*{(a) Fog.}
\end{minipage}
\begin{minipage}{1\linewidth}
\centering
\subfloat{\includegraphics[width=.32\linewidth]{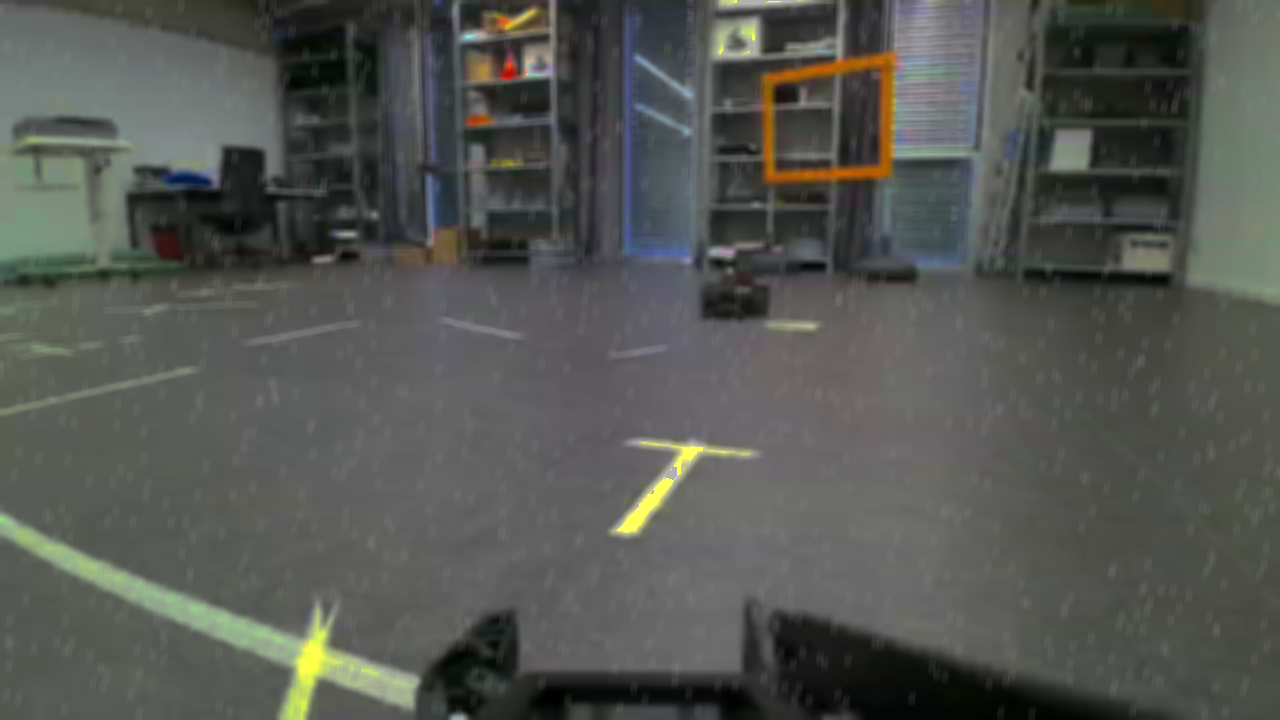}}
\subfloat{\includegraphics[width=.32\linewidth]{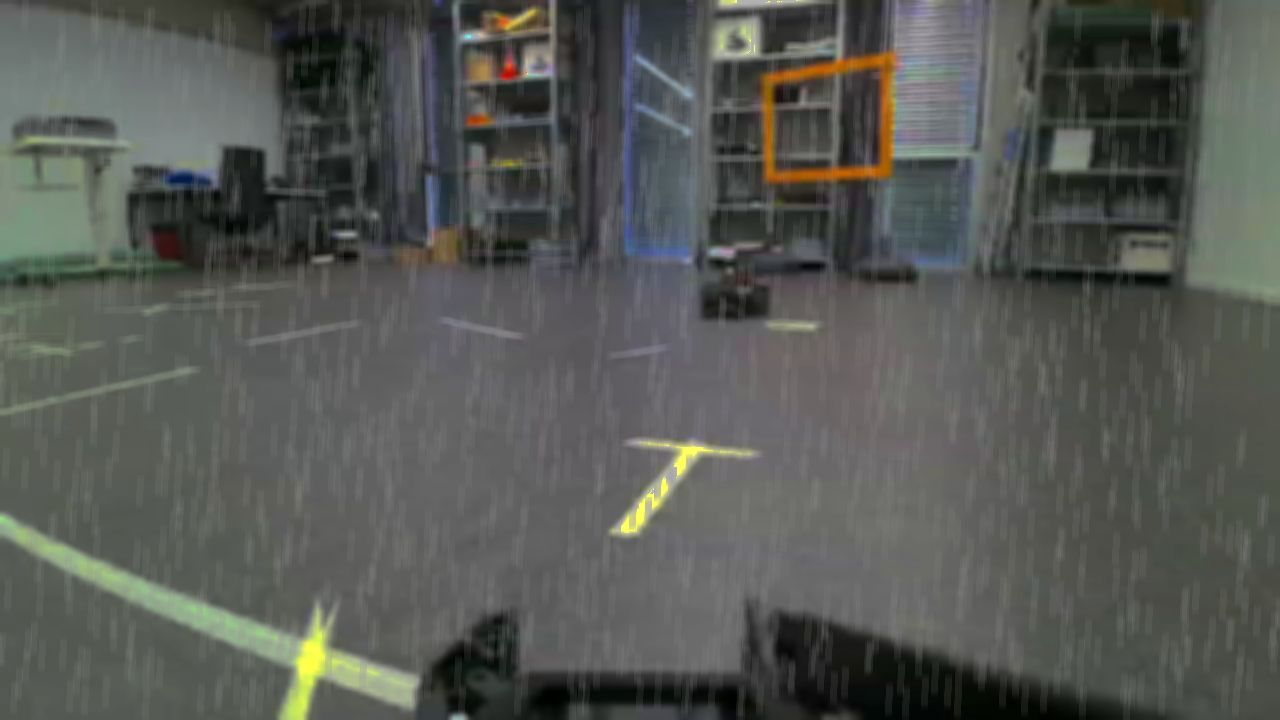}}
\subfloat{\includegraphics[width=.32\linewidth]{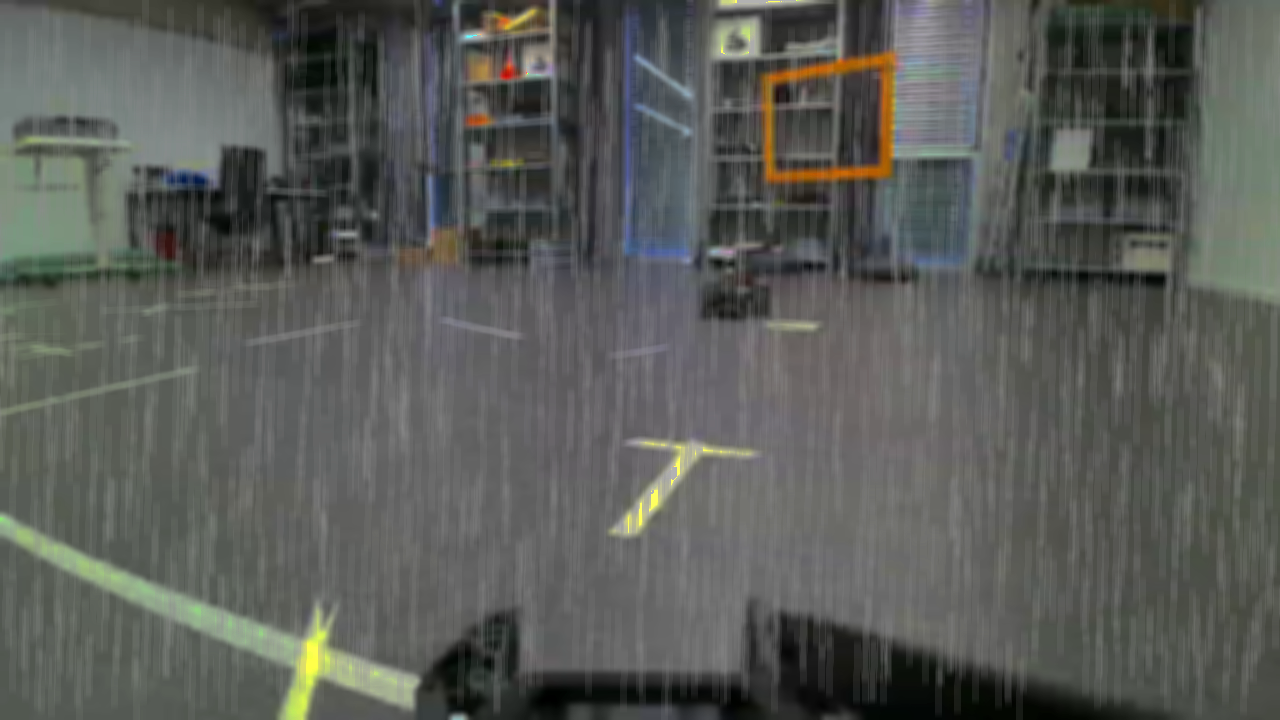}}
\caption*{(b) Rain.}
\end{minipage}
\begin{minipage}{1\linewidth}
\centering
\subfloat{\includegraphics[width=.32\linewidth]{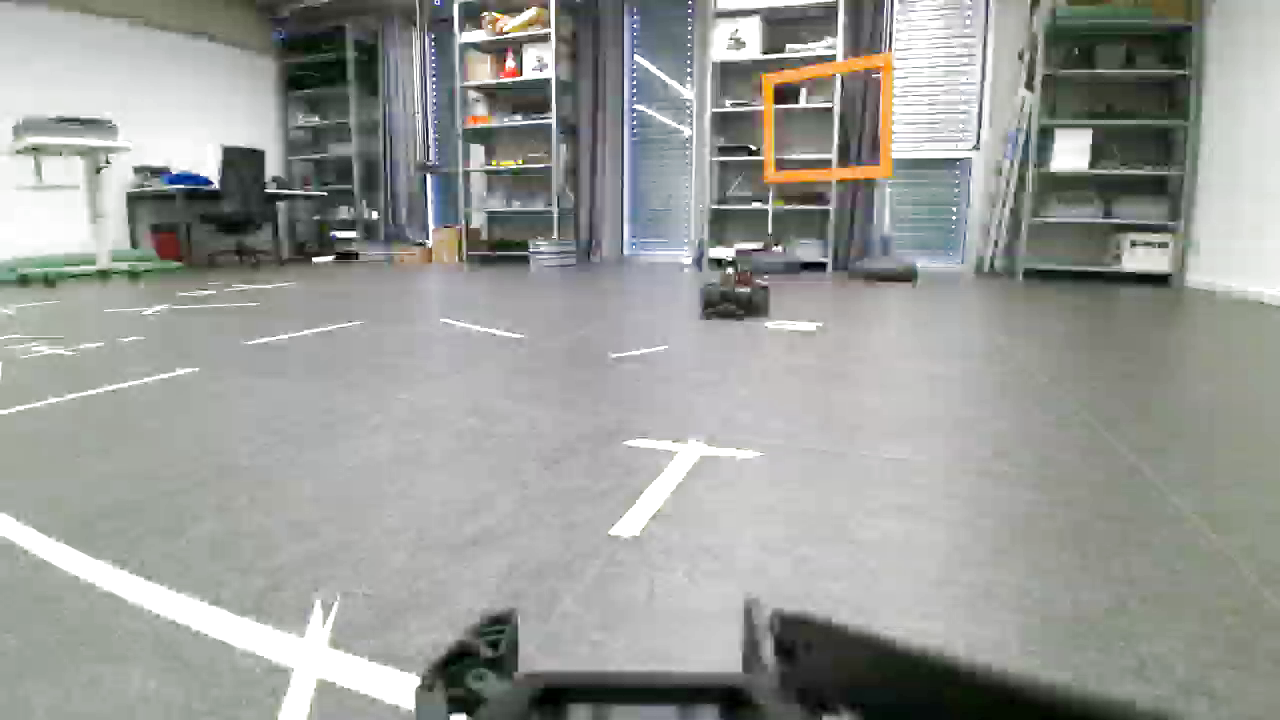}}
\subfloat{\includegraphics[width=.32\linewidth]{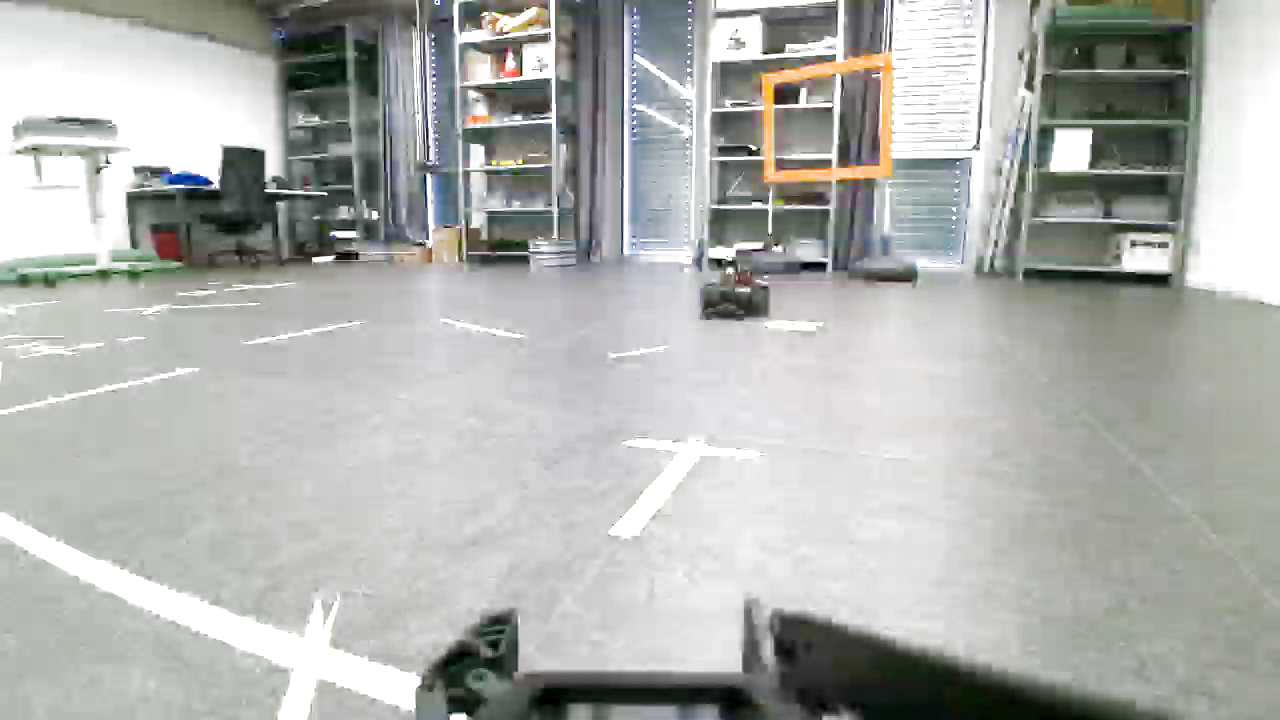}}
\subfloat{\includegraphics[width=.32\linewidth]{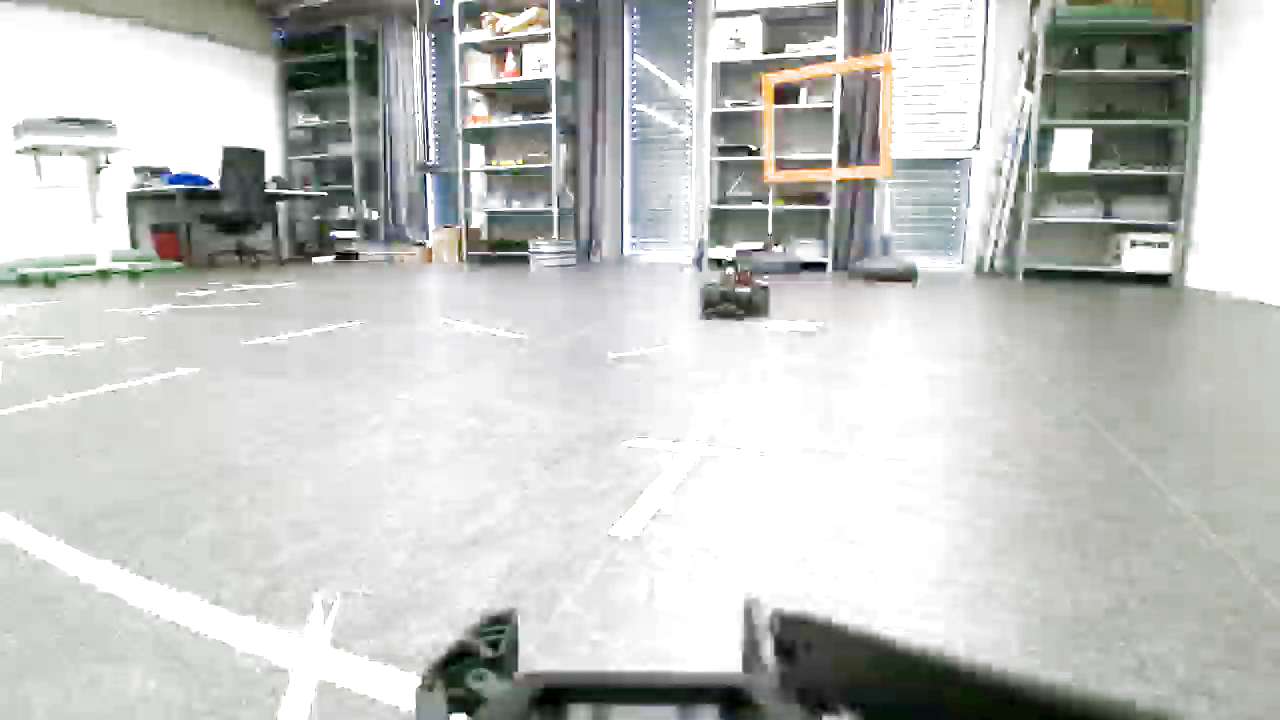}}
\caption*{(c) Brightness.}
\end{minipage}\quad
\begin{minipage}{1\linewidth}
\centering
\subfloat{\includegraphics[width=.32\linewidth]{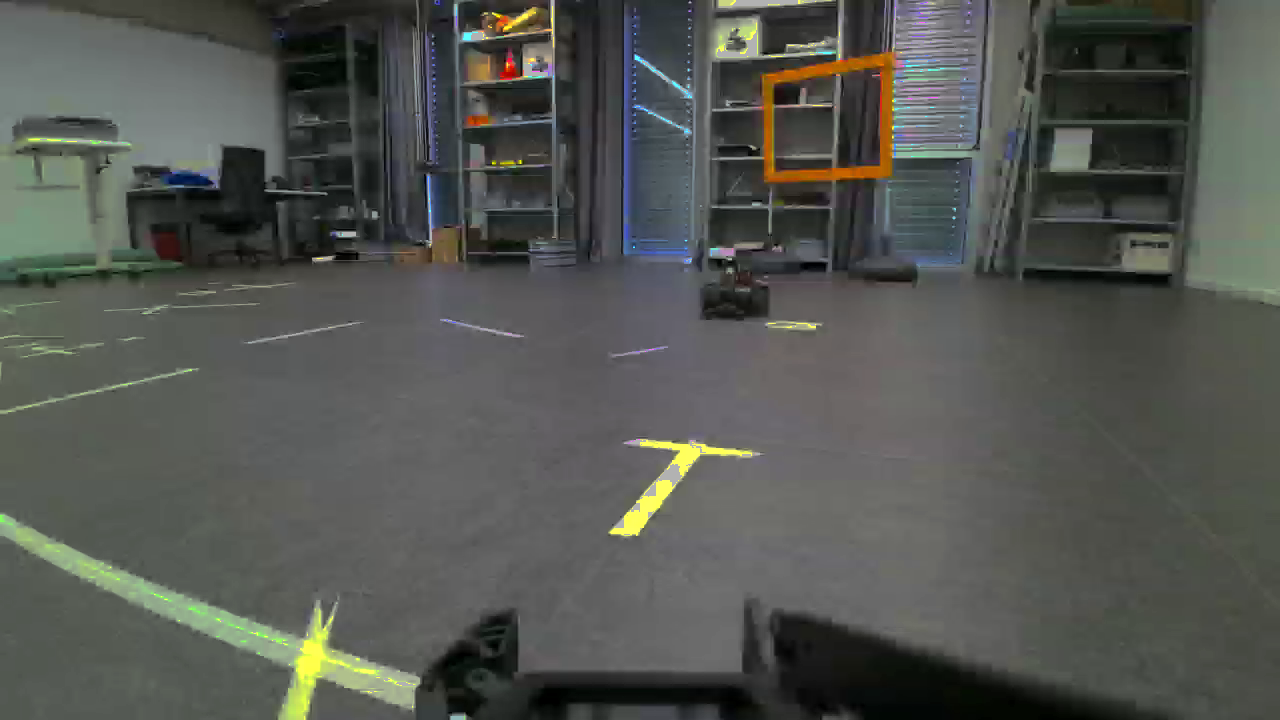}}
\subfloat{\includegraphics[width=.32\linewidth]{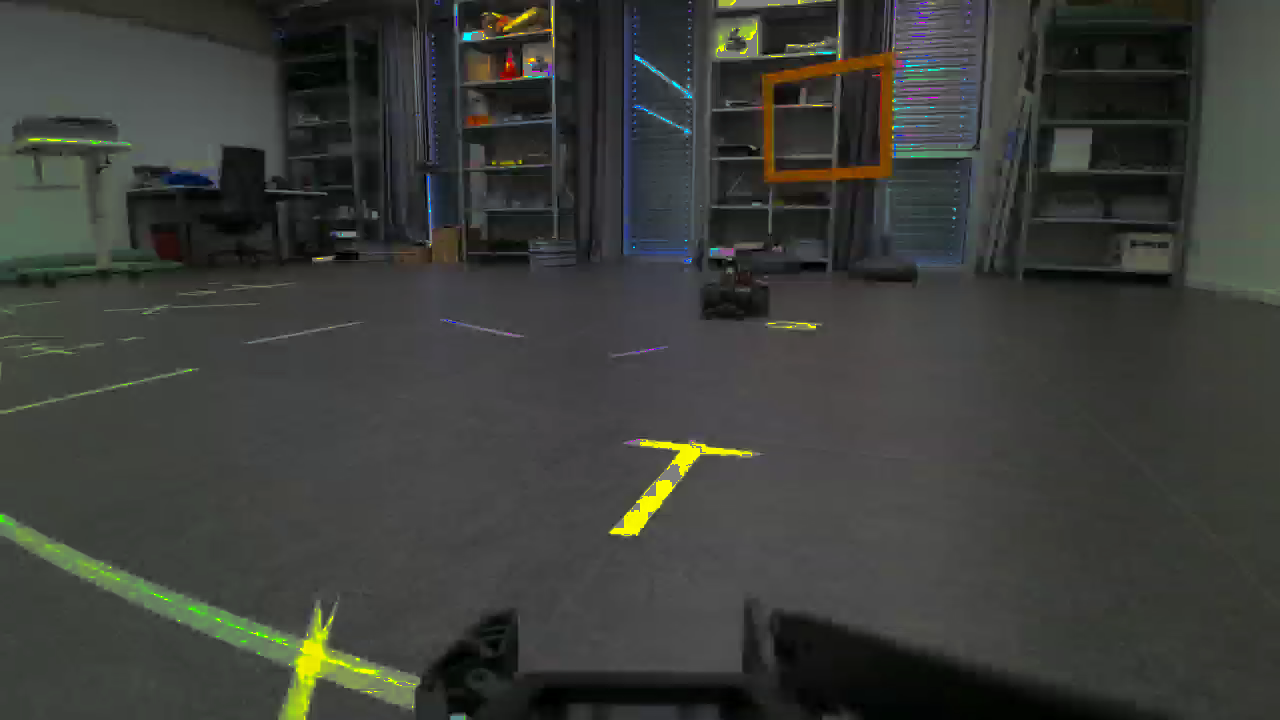}}
\subfloat{\includegraphics[width=.32\linewidth]{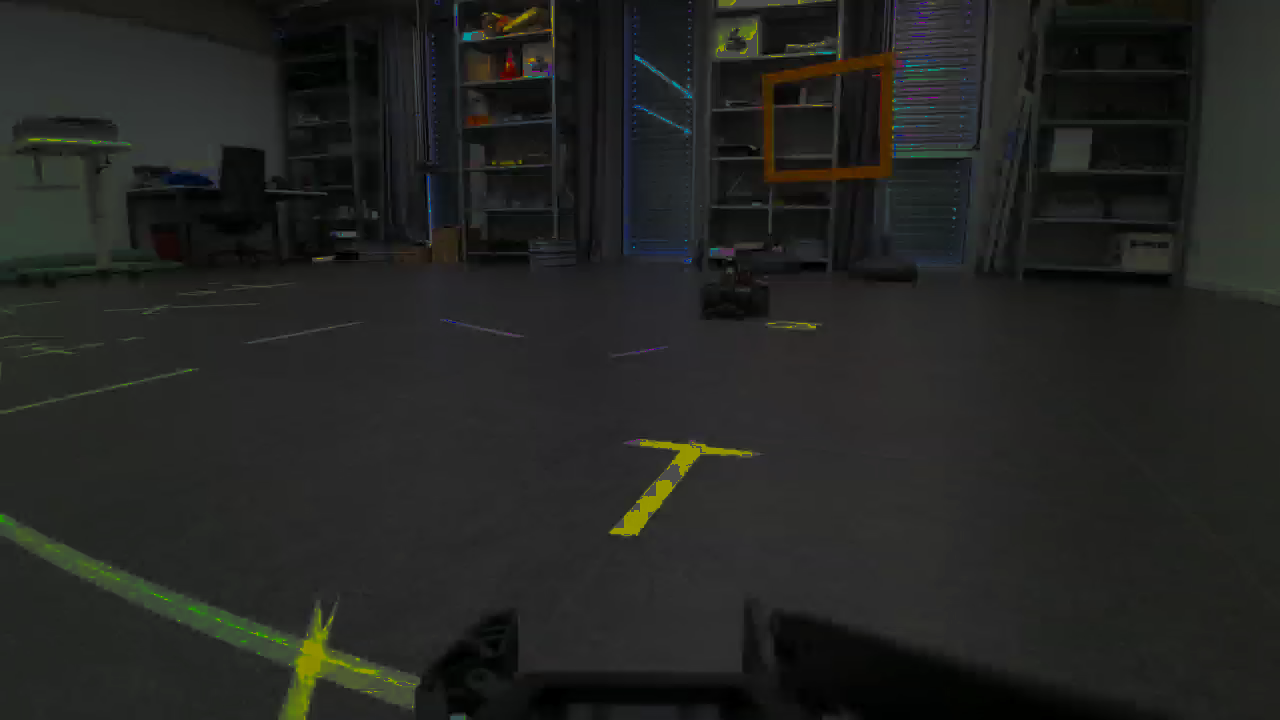}}
\caption*{(d) Darkness.}
\end{minipage}
\caption{Filters simulating environmental conditions.}
\label{fig:data_aug}
\end{figure}

Raw data from cameras are split into frames using the OpenCV library.\footnote{\url{https://docs.opencv.org/4.x/index.html}.} The environmental factors are simulated at the frame level by the Automold library for data augmentation.\footnote{\url{https://github.com/UjjwalSaxena/Automold--Road-Augmentation-Library}.} 
We use the library to simulate the following effects: 
\begin{itemize}
\item \emph{fog} with coefficient $0.3$, $0.5$ and $0.7$ (Fig.~\ref{fig:data_aug}.a);
\item \emph{rain} of type \emph{drizzle}, \emph{heavy}, and \emph{torrential} (Fig.~\ref{fig:data_aug}.b);
\item \emph{bright} with coefficient $0.3$, $0.5$, $0.7$, and $0.9$ (Fig.~\ref{fig:data_aug}.c);
\item \emph{dark} with coefficient $0.3$, $0.5$, $0.7$, $0.9$ (Fig.~\ref{fig:data_aug}.d).
\end{itemize}
We refer to the unfiltered images and sequences as the \emph{original} ones.

For obstacle detection, original and augmented frames are processed by YOLO (\emph{You Only Look Once}).\footnote{\url{https://github.com/ultralytics/yolov5}.} YOLO is a state-of-the-art object detection architecture, widely used for its real-time performances, generalization and adaptability capabilities. 

\begin{figure}[htp!]%
\centering
\subfloat[]{\includegraphics[height=3cm]{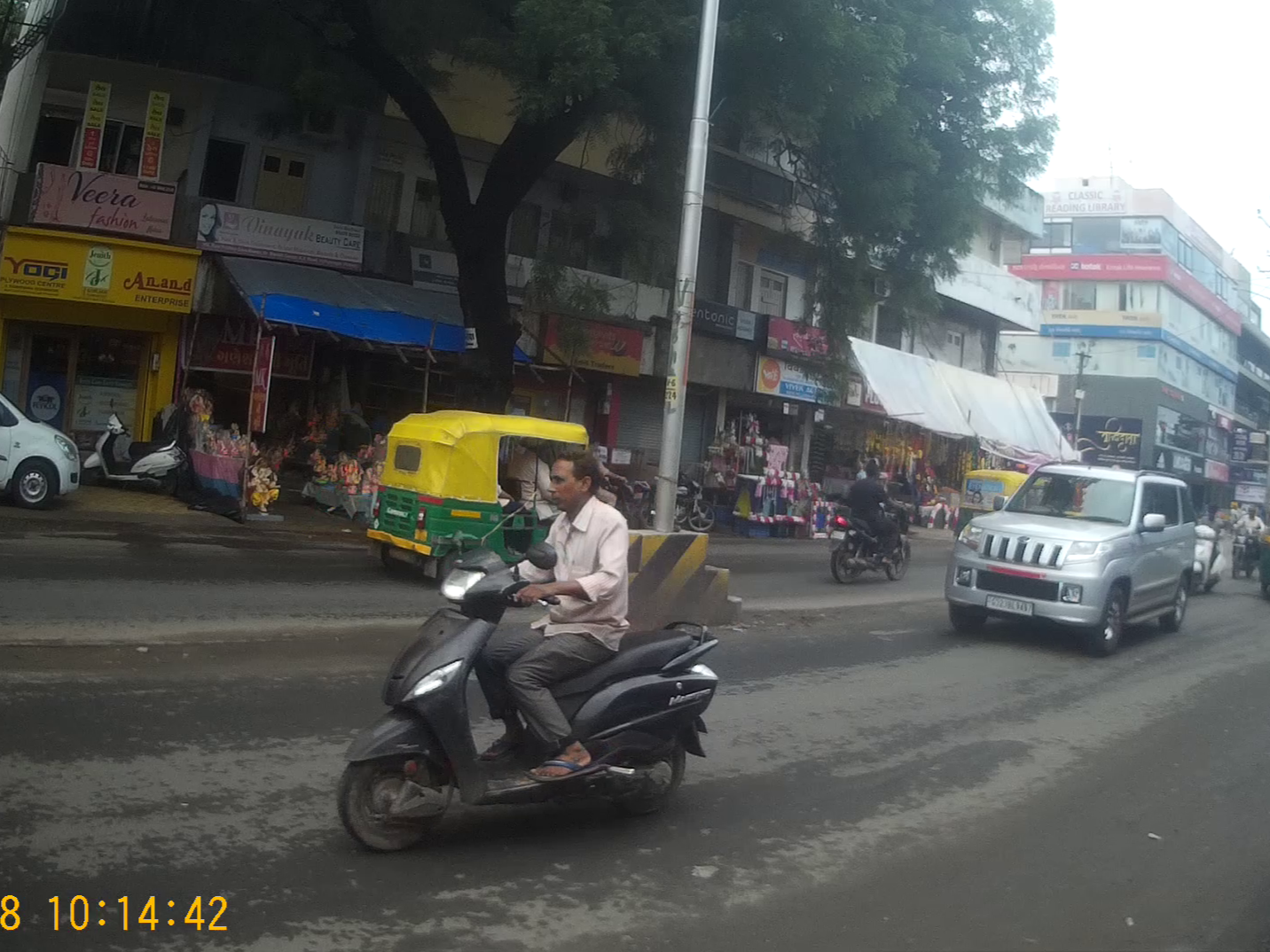}}
\hfill
\subfloat[]{\includegraphics[height=3cm]{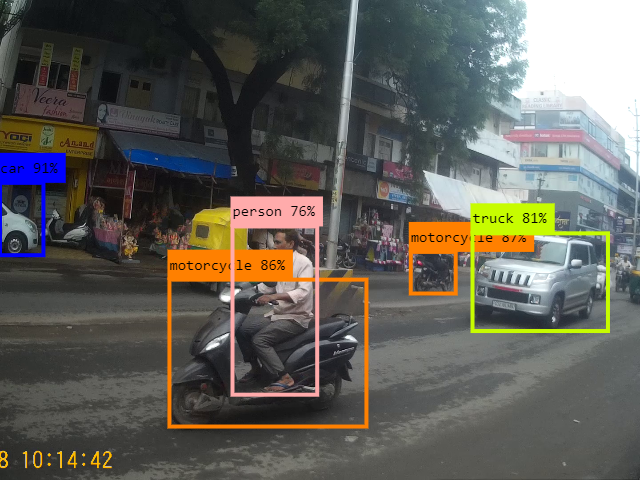}}
\caption{A real-world road traffic scenario \cite{brahmbhatt2022dataset} and the corresponding object detection provided by YOLO trained on the COCO dataset \cite{cocodataset}.}
\label{fig:YOLO}%
\end{figure}

YOLO's pre-trained deep neural networks are based on the COCO image dataset \cite{cocodataset}, whose classes are, among others, person, bicycle, car, motorcycle, bus, train, and truck (see, e.g., Fig.~\ref{fig:YOLO}). Note that, together with the recognition of the object class, YOLO also provides the coordinates of the bounding box for the recognised object (or objects) within the frame. The generic model YOLO v5, used in \cite{grigioni2024safe} for obstacle detection from the cameras videos, does not have a specific class for the RM device. Yet, it is able to consistently recognise it under the \emph{motorcycle} class. As an alternative to such a pre-trained model, we consider the possibility of a fine tuning. This is achieved by means of a dataset of RM images\footnote{\url{https://universe.roboflow.com/godwyll-aikins/robomaster-i5ydd}.} within the YOLO v8 architecture. In the following, we refer to the pre-trained architecture as Y5 and to the fine-tuned one as Y8. 

To guarantee redundancy and diversity during the information fusion, we consider both Y5 and Y8. Both models are therefore applied to all the frames of all the experiments to identify the bounding boxes of the obstacle (identified as a motorcycle by Y5 and as a RM by Y8). Therefore, there are a total of four data flows for information fusion: two processed by the RM$_{\mathrm{d}}$ camera using two different obstacle detection models, and, processed in the same way, another two by the RM$_{\mathrm{w}}$ camera. It is important to underline that in this approach redundancy and sensory diversity are obtained through software thanks to the use of different ML models. 

From the bounding boxes, we eventually compute the distance of the object \cite{distance_measurment}. 
In case of missing data, due to the absence of obstacles or sensor malfunction, a \emph{reference} maximum distance value is imputed. This equates to the maximum detectable distance of interest, which in the case of our tests is 4 m.
This strategy correctly identifies situations in which the obstacle has never been seen, or is no longer visible because it has exited the visual field of interest. 



\section{Experimental results}\label{sec:experiments}
In this section, we present the results of our experimental analysis. In total, 12 of the 15 sequences are used, six for scenario A and six for scenario B. The remaining ones, recorded in scenario B, are omitted to have the same number of tests per scenario.

As discussed in the previous section, we apply relevant filters to the video sequences to simulate environmental conditions. The original, unperturbed, video sequences are also considered for comparison. For each sequence, we consider four signals generated by the two cameras on the RM$_{\mathrm{w}}$ and the RM$_{\mathrm{d}}$, whose frames are processed by both Y5 and Y8. Those four signals can be fused by a trivial sensor fusion approach involving the arithmetic average of the position measurements for a particular frame as discussed in Sect.~\ref{sec:fusion}. We denote such a procedure as \emph{A-Fusion}. As an alternative approach, denoted in the following as \emph{W-Fusion}, we consider an arithmetic fusion of the distance measurements that removes the signal that are not based on actual measurements but on imputations. If all the four measures are based on an imputation, we just consider the \emph{reference} maximum distance.

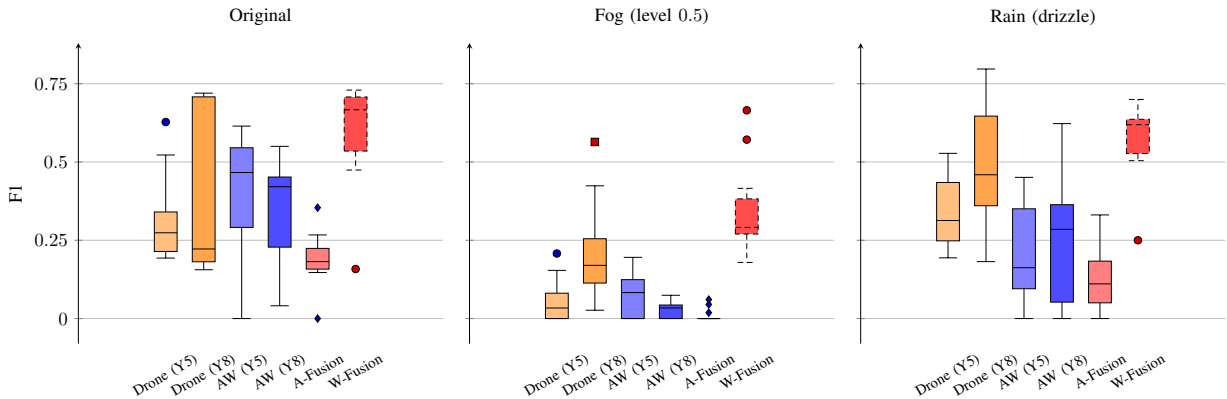
\begin{figure*}[htp!]
\begin{tikzpicture}[scale=0.7]
\pgfplotstableread[col sep=comma]{data.csv}\csvdata
\pgfplotstabletranspose\datatransposed{\csvdata} 
\begin{axis}[
boxplot/draw direction = y,
x axis line style = {opacity=0},
axis x line* = bottom,
axis y line = left,
enlarge y limits,
ymajorgrids,
ymin=0,
ymax=0.8,
title={Original},
xtick = {1,2,3,4,5,6},
xticklabel style = {align=center, font=\footnotesize, rotate=30},
xticklabels = {Drone (Y5), Drone (Y8), AW (Y5), AW (Y8), A-Fusion, W-Fusion},
xtick style = {draw=none}, 
ylabel = {F1},
ytick = {0, 0.25, 0.5, 0.75},
boxplot/variable width,
boxplot/box extend=3
]
\addplot+[boxplot, fill, color=orange!50, draw=black] table[y index=1] {\datatransposed};
\addplot+[boxplot, fill, color=orange!70, draw=black] table[y index=2] {\datatransposed};
\addplot+[boxplot, fill, color=blue!50, draw=black] table[y index=3] {\datatransposed};
\addplot+[boxplot, fill, color=blue!70, draw=black] table[y index=4] {\datatransposed};
\addplot+[boxplot, fill, color=red!50, draw=black] table[y index=5] {\datatransposed};
\addplot+[boxplot, fill, color=red!70, draw=black] table[y index=6] {\datatransposed};
\end{axis}
\end{tikzpicture}
\begin{tikzpicture}[scale=0.7]
\pgfplotstableread[col sep=comma]{datafog2.csv}\csvdatafog
\pgfplotstabletranspose\datatransposed{\csvdatafog} 
\begin{axis}[
boxplot/draw direction = y,
x axis line style = {opacity=0},
axis x line* = bottom,
axis y line = left,
enlarge y limits,
ymin=0,
ymax=0.8,
ymajorgrids,
xtick = {1,2,3,4,5,6},
xticklabel style = {align=center, font=\footnotesize, rotate=30},
xticklabels = {Drone (Y5), Drone (Y8), AW (Y5), AW (Y8), A-Fusion, W-Fusion},
title={Fog (level $0.5$)},
xtick style = {draw=none}, 
ytick = {0, 0.25, 0.5, 0.75},
yticklabels = {},
boxplot/variable width,
boxplot/box extend=3
]
\addplot+[boxplot, fill, color=orange!50, draw=black] table[y index=1] {\datatransposed};
\addplot+[boxplot, fill, color=orange!70, draw=black] table[y index=2] {\datatransposed};
\addplot+[boxplot, fill, color=blue!50, draw=black] table[y index=3] {\datatransposed};
\addplot+[boxplot, fill, color=blue!70, draw=black] table[y index=4] {\datatransposed};
\addplot+[boxplot, fill, color=red!50, draw=black] table[y index=5] {\datatransposed};
\addplot+[boxplot, fill, color=red!70, draw=black] table[y index=6] {\datatransposed};
\end{axis}
\end{tikzpicture}
\begin{tikzpicture}[scale=0.7]
\pgfplotstableread[col sep=comma]{datarain1.csv}\csvdatarain
\pgfplotstabletranspose\datatransposed{\csvdatarain} 
\begin{axis}[
boxplot/draw direction = y,
x axis line style = {opacity=0},
axis x line* = bottom,
axis y line = left,
enlarge y limits,
ymin=0,
ymax=0.8,
ymajorgrids,
xtick = {1,2,3,4,5,6},
xticklabel style = {align=center, font=\footnotesize, rotate=30},
xticklabels = {Drone (Y5), Drone (Y8), AW (Y5), AW (Y8), A-Fusion, W-Fusion},
title={Rain (drizzle)},
xtick style = {draw=none}, 
ytick = {0, 0.25, 0.5, 0.75},
yticklabels = {},
boxplot/variable width,
boxplot/box extend=3
]
\addplot+[boxplot, fill, color=orange!50, draw=black] table[y index=1] {\datatransposed};
\addplot+[boxplot, fill, color=orange!70, draw=black] table[y index=2] {\datatransposed};
\addplot+[boxplot, fill, color=blue!50, draw=black] table[y index=3] {\datatransposed};
\addplot+[boxplot, fill, color=blue!70, draw=black] table[y index=4] {\datatransposed};
\addplot+[boxplot, fill, color=red!50, draw=black] table[y index=5] {\datatransposed};
\addplot+[boxplot, fill, color=red!70, draw=black] table[y index=6] {\datatransposed};
\end{axis}
\end{tikzpicture}
\caption{F1 scores boxplots for road-crossing decision.}\label{fig:f1}
\end{figure*}

As performance descriptors, we consider the percentage of frames where the target has been recognized by YOLO (Tab.~\ref{tab:missing}), the \textit{Root Mean Square Error} (RMSE) of the DF based on the sensors measurements, or the sensors fusion procedures, compared against the ground-truth measurements provided by the tracker (Tab.~\ref{tab:rmse}). We similarly proceed for the road-crossing decisions, based on the DF value compared against the threshold levels, by reporting accuracies (Tab.~\ref{tab:accuracy}) and F1 measures (Fig.~\ref{fig:f1}). For the quantitative descriptors we provide mean values over the different sequences and, in parentheses, the standard deviations. Those data are reported separately for the two scenarios. The only exception is the F1 score, for which, for the sake of simplicity, we prefer to depict the aggregated results over the two scenarios as boxplots.

Regarding the frame recognition rates in Tab.~\ref{tab:missing}, the filters simulating adverse weather conditions typically induce rates that are lower than those in the original sequences. In extreme conditions (i.e., fog with coefficient $0.7$ and heavy rain) those rates are close to zero, and this makes the performance very poor. The corresponding results are not reported for the sake of space. 

As expected, the Y8 model results to work better than the Y5 model, especially for highly adverse conditions (e.g., \textit{heavy rain}). However, there is a clear difference in performance between the data from RM$_{\mathrm{d}}$ (better one) and RM$_{\mathrm{w}}$. Indeed, the best performance is achieved using the videos recorded by the RM$_{\mathrm{d}}$, probably due to the fact that in these shots the RM$_{\mathrm{d}}$ has poses similar to those of the dataset used to fine tune the Y8 model.

Unexpectedly, the brightness and darkness filters with low coefficient levels (i.e., 0.3 and 0.5) induced higher recognition rates. In practice, such filters act as an image preprocessing improving the perfomance of obstacle recognition. Probably this is due to the blurring of background elements, so that the RM$_{\mathrm{c}}$ is better recognisable from the YOLO models.
For \emph{bright} and \emph{dark} filters, the rates are quite similar for the different values of the coefficients, and we therefore report only the data for a single coefficient for each filter.
We also measure the difference between the distances induced by the original frames and the filtered ones. Apart from filters leading to almost zero recognition rates, we observe low RMSE values ($\leq 0.2\mathrm{m}$). This seems to suggest that, rather than leading to very inaccurate measures, the main negative effect induced by the exogenous factors is to prevent the recognition of the objects.


\begin{table*}[htp!]
{\centering
\footnotesize
\begin{tabular}{lllp{.5cm}p{.6cm}p{.5cm}p{.6cm}p{.5cm}p{.6cm}p{.5cm}p{.6cm}p{.5cm}p{.6cm}p{.5cm}p{.6cm}p{.5cm}p{.6cm}}
\toprule
\multirow{2}{*}{Sc.}&\multirow{2}{*}{Cam}&\multirow{2}{*}{YOLO}&\multicolumn{2}{c}{\multirow{2}{*}{Original}}&\multicolumn{4}{c}{Fog}&\multicolumn{4}{c}{Rain}&\multicolumn{2}{c}{Bright}&\multicolumn{2}{c}{Dark}\\

&&&&&\multicolumn{2}{c}{$0.3$}&\multicolumn{2}{c}{$0.5$}&\multicolumn{2}{c}{drizzle}&\multicolumn{2}{c}{heavy}&\multicolumn{2}{c}{$0.5$}&\multicolumn{2}{c}{$0.5$}\\

\midrule
\multirow{4}{*}{A}&\multirow{2}{*}{Drone}&Y5&20.52&(4.04)&21.11&(4.94)&7.61&(4.31)&27.33&(4.03)&0.82&(0.72)& 18.10 & (3.97) &17.28&(4.06)\\

&&Y8&73.41&(9.21)&52.30&(10.37)&28.16&(9.25)&73.11&(9.52)&69.00&(7.99)&75.32&(8.49)&76.10&(7.95)\\

&\multirow{2}{*}{AW}&Y5&35.96&(18.16)&31.53&(15.96)&6.44&(6.34)&25.82&(13.24)&0.20&(0.50)&31.73&(16.56)&36.61&(5.87)\\

&&Y8&41.63&(5.45)&8.12&(9.92)&6.13&(1.57)&29.30&(14.07)&24.83&(13.80)&46.56&(11.28)&52.31&(10.90)\\

\midrule
\multirow{4}{*}{B}&\multirow{2}{*}{Drone}&Y5&38.03&(12.22)&35.59&(10.75)&11.12&(4.33)&38.27&(10.80)&0.16&(0.39)&42.23&(12.69)&43.42&(12.74)\\

&&Y8&47.39&(8.87)&52.02&(4.66)&19.09&(6.22)&40.53&(3.56)&26.16&(2.61)&81.42&(6.95)&81.17&(6.10)\\

&\multirow{2}{*}{AW}&Y5&31.86&(10.03)&27.10&(14.29)&18.53&(10.54)&15.53&(5.63)&0.20&(0.32)&27.41&(11.31)&26.66&(11.44)\\

&&Y8&25.51&(5.45)&12.44&(4.31)&4.59&(1.62)&15.02&(5.31)&14.16&(5.12)&58.31&(5.95)&64.70&(9.63)\\
\bottomrule
\end{tabular}}
\caption{Means and deviations of the frame recognition rates of YOLO pre-trained (Y5) and fine-tuned (Y8).}\label{tab:missing}
\end{table*}

\begin{table*}[htp!]
{\centering
\footnotesize
\begin{tabular}{lllp{.5cm}p{.6cm}p{.5cm}p{.6cm}p{.5cm}p{.6cm}p{.5cm}p{.6cm}p{.5cm}p{.6cm}p{.5cm}p{.6cm}p{.5cm}p{.6cm}}
\toprule
\multirow{2}{*}{Sc.}&\multirow{2}{*}{Cam}&\multirow{2}{*}{YOLO}&\multicolumn{2}{c}{\multirow{2}{*}{Original}}&\multicolumn{4}{c}{Fog}&\multicolumn{4}{c}{Rain}&\multicolumn{2}{c}{Bright}&\multicolumn{2}{c}{Dark}\\
&&&&&\multicolumn{2}{c}{$0.3$}&\multicolumn{2}{c}{$0.5$}&\multicolumn{2}{c}{drizzle}&\multicolumn{2}{c}{heavy}&\multicolumn{2}{c}{$0.5$}&\multicolumn{2}{c}{$0.5$}\\
\midrule
\multirow{6}{*}{A}&\multirow{2}{*}{Drone}&Y5&1.44&(0.40)&1.39&(0.38)&1.52&(0.45)&1.46&(0.38)&1.55&(0.47)&1.40&(0.41)&1.38&(0.34)\\

&&Y8&1.15&(0.45)&1.33&(0.39)&1.54&(0.37)&1.13&(0.42)&1.26&(0.46)&1.15&(0.42)&1.15&(0.43)\\

&\multirow{3}{*}{AW}&Y5&1.35&(0.35)&1.40&(0.38)&1.55&(0.45)&1.44&(0.39)&1.55&(0.47)&1.37&(0.36)&1.35&(0.35)\\
&&Y8&1.43&(0.44)&1.53&(0.44)&1.54&(0.46)&1.46&(0.40)&1.49&(0.42)&1.41&(0.43)&1.50&(0.42)\\
\cmidrule{2-17}
&\multicolumn{2}{l}{A-Fusion}&1.22&(0.46)&1.29&(0.44)&1.40&(0.44)&1.23&(0.41)&1.32&(0.43)&1.20&(0.45)&1.25&(0.43)\\
&\multicolumn{2}{l}{W-Fusion}&1.09&(0.41)&1.15&(0.41)&1.49&(0.37)&1.17&(0.40)&1.22&(0.46)&1.07&(0.41)&1.07&(0.35)\\
\midrule
\multirow{6}{*}{B}&\multirow{2}{*}{Drone}&Y5&1.26&(0.32)&1.25&(0.29)&1.16&(0.30)&1.28&(0.35)&1.13 &(0.34)&1.27&(0.35)&1.28&(0.27)\\

&&Y8&1.06&(0.29)&1.03&(0.19)&1.39&(0.28)&1.02 &(0.27)&1.10&(0.33)&0.90&(0.16)&1.11&(0.29)\\

&\multirow{2}{*}{AW}&Y5&1.08&(0.22)&1.20&(0.19)&1.23 &(0.22)&1.11&(0.31)&1.13&(0.34)&1.17&(0.25)&1.07&(0.24)\\

&&Y8&1.01&(0.29)&1.07&(0.31)&1.12&(0.33)&1.05&(0.30)&1.06&(0.30)&0.93&(0.28)&1.09&(0.31)\\

\cmidrule{2-17}
&\multicolumn{2}{l}{A-Fusion}&0.88&(0.27)&0.88&(0.28)&1.06&(0.34)&0.93&(0.29)&1.01&(0.31)&0.82&(0.28)&0.97&(0.29)\\

&\multicolumn{2}{l}{W-Fusion}&1.00&(0.13)&1.12&(0.18)&1.36&(0.22)&1.13&(0.24)&1.01&(0.29)&1.01&(0.23)&1.12&(0.16)\\

\bottomrule
\end{tabular}}
\caption{Means and deviations of the RMSE for the danger function.}\label{tab:rmse}
\end{table*}

\begin{table*}[htp!]
{\centering
\footnotesize
\begin{tabular}{lllp{.5cm}p{.6cm}p{.5cm}p{.6cm}p{.5cm}p{.6cm}p{.5cm}p{.6cm}p{.5cm}p{.6cm}p{.5cm}p{.6cm}p{.5cm}p{.6cm}}
\toprule
\multirow{2}{*}{Sc.}&\multirow{2}{*}{Cam}&\multirow{2}{*}{YOLO}&\multicolumn{2}{c}{\multirow{2}{*}{Original}}&\multicolumn{4}{c}{Fog}&\multicolumn{4}{c}{Rain}&\multicolumn{2}{c}{Bright}&\multicolumn{2}{c}{Dark}\\
&&&&&\multicolumn{2}{c}{$0.3$}&\multicolumn{2}{c}{$0.5$}&\multicolumn{2}{c}{drizzle}&\multicolumn{2}{c}{heavy}&\multicolumn{2}{c}{$0.5$}&\multicolumn{2}{c}{$0.5$}\\
\midrule
\multirow{6}{*}{A}&\multirow{2}{*}{Drone}&Y5&44.64&(7.98)&48.45&(6.35)&36.63&(7.28)&46.04&(8.21)&35.55&(8.15)& 44.74& (7.04)& 49.75& (5.47)\\
&&Y8&69.55&(4.55)&59.06&(13.94)&46.23&(13.90)&69.56&(5.59)&63.76&(4.73)& 73.92& (6.12)& 68.90& (4.86)\\
&\multirow{3}{*}{AW}&Y5&57.29&(6.94)&51.85&(7.29)&36.65&(8.43)&49.10&(5.37)&35.55&(8.15)& 54.29& (7.05)& 58.04& (6.72)\\
&&Y8&48.61&(2.66)&37.77&(7.24)&37.08&(7.75)&46.01&(8.91)&41.59&(4.29)& 48.06& (3.92)& 43.26& (5.51)\\
\cmidrule{2-17}
&\multicolumn{2}{l}{A-Fusion}&43.43&(3.92)&43.58&(5.12)&35.55&(8.15)&42.06&(5.10)&35.55&(8.15)& 44.53& (4.25)& 44.17& (4.52)\\
&\multicolumn{2}{l}{W-Fusion}&69.37&(5.10)&63.53&(10.09)&53.44&(13.28)&64.39&(5.20)&62.19&(5.36)& 71.39& (4.95)& 70.09& (2.53)\\
\midrule
\multirow{6}{*}{B}&\multirow{2}{*}{Drone}&Y5&67.95&(11.02)&68.75&(9.60)&62.62&(18.12)&68.04&(9.43)&62.34&(18.89)& 70.94& (6.69)& 66.01& (11.26)\\
&&Y8&63.22&(15.09)&65.23&(12.42)&62.26&(16.10)&66.45&(12.49)&63.09&(17.41)& 75.17& (5.25)& 62.90& (16.70)\\
&\multirow{2}{*}{AW}&Y5&68.49&(11.52)&65.61&(14.55)&63.16&(15.88)&64.50&(16.78)&62.34&(18.89)& 64.01& (13.72)& 68.65& (12.12)\\
&&Y8&72.18&(10.20)&66.26&(15.51)&62.73&(18.68)&67.73&(13.11)&66.99&(14.75)& 71.90& (8.53)& 65.19& (15.54)\\
\cmidrule{2-17}
&\multicolumn{2}{l}{A-Fusion}&64.49&(14.93)&66.32&(14.17)&62.35&(18.53)&64.02&(16.41)&62.34&(18.89)& 66.84& (12.71)& 64.07& (15.74)\\
&\multicolumn{2}{l}{W-Fusion}&72.30&(5.54)&70.98&(5.89)&63.60&(13.97)&72.16&(6.67)&68.58&(11.49)& 74.49& (4.48)& 70.03& (9.22)\\
\bottomrule
\end{tabular}}
\caption{Means and deviations accuracy for road-crossing decision.}\label{tab:accuracy}
\end{table*}

Regarding the RMSE and accuracy values in Tab.~\ref{tab:rmse} and Tab.~\ref{tab:accuracy}, similar observations to Tab.~\ref{tab:missing} can be obtained. Adverse environmental filters generally lead, with a few minor anomalies, to a deterioration in performance with the exception of the \textit{bright} and \textit{dark} filters, which keep them constant or even improve them.
The two fusion \emph{W-Fusion} and \emph{A-Fusion} do not differ significantly in terms of RMSE. On the other hand, \emph{W-Fusion} is definitely better considering the accuracy values in Tab.~\ref{tab:accuracy}.
This means that although the two fusion methods do not differ much from each other in terms of risk evaluation, the \emph{W-Fusion} method is more precise for road-crossings decisions.
For a deeper analysis of the quality of the road-crossing decisions provided by our system, we also consider the F1 performance. The boxplots in Fig.~\ref{fig:f1} for the original sequences and for two filters clearly show the advantages of the \emph{W-Fusion} procedure we proposed with respect to a trivial arithmetic average (\emph{A-Fusion}). Notably, the performance of our fusion procedure is also outperforming that of the individual sensors, this advocating the advantages of adding redundancy and diversity to the sensor system equipment.

In conclusion, both generic Y5 and fine-tuned Y8 are negatively influenced by adverse weather, with the exception of the \textit{bright} and \textit{dark} filter. However, Y8 appears to be more robust against environmental disturbances. The new \emph{W-Fusion} outperforms the simpler \emph{A-Fusion} and all single sensors in terms of accuracy and F1 performance for the road-crossing decision.

\section{Conclusions}\label{sec:conc}
Our study delved into the crucial aspect of road-crossing decisions by autonomous wheelchairs, supported by flying drones, in the face of various environmental factors. By focusing on the refinement of obstacle detection performance through artificial vision and its evaluation against outdoor conditions like fog, rain, brightness and darkness, we aimed to quantify the impact of these factors on tracking accuracy.

Adverse environmental conditions were found to significantly challenge video tracking performance. However, our approach facilitated a systematic assessment of these effects within the specific operational context. By leveraging available laboratory datasets and employing tailored software filters, we demonstrated the feasibility of evaluating video tracking robustness against environmental variables in reference scenarios. Our findings underscore the importance of proactively addressing environmental challenges in autonomous wheelchair navigation systems. By identifying instances where tracking accuracy falls below acceptable thresholds, our approach enables the issuance of timely warnings and the potential reconfiguration of sensor priorities to enhance overall safety. Furthermore, the incorporation of supplementary sensors such as light and rain sensors offers additional layers of detection for critical situations.

Our study contributes to the ongoing efforts in advancing the trustworthiness of autonomous wheelchair technology. By systematically evaluating and mitigating the impact of environmental factors on road-crossing decisions, we pave the way for more robust and dependable autonomous navigation systems, ultimately enhancing the mobility and independence of individuals with mobility impairments. 

The approach presented in this study can be generalized and therefore applied to a larger class of cooperative autonomous robots and self-driving vehicles in order to test their robustness against environmental factors. In the future, we plan to extend the experimentation in multiple operating conditions and scenarios, also including other sensors that can be affected by different disturbances.

\addtolength{\textheight}{-12cm}
\section*{ACKNOWLEDGMENT}
This work was supported by the Swiss State Secretariat for Education, Research and Innovation (SERI) under contract no. 22.00291 (REXASI-PRO project).
The project has been selected within the European Union’s Horizon Europe research and innovation programme under grant agreement ID: 101070028 (call HORIZON-CL4-2021-HUMAN-01-01).
Views and opinions expressed are however those of the authors only and do not necessarily reflect those of the funding agencies, which cannot be held responsible for them.




\bibliographystyle{splncs04}
\bibliography{bibliography}
\end{document}